\documentclass[final,5p,times,twocolumn]{elsarticle}

\biboptions{sort&compress}
\usepackage{amssymb}
\usepackage{amsthm}
\usepackage{amsmath}
\usepackage{multicol}
\usepackage{multirow}
\usepackage{siunitx}
\usepackage{xparse}
\usepackage{booktabs}
\usepackage{graphicx}
\usepackage{svg}
\usepackage{enumitem}
\usepackage{tabularx} 

\journal{TO SPECIFY}

\begin{document}
\begin{frontmatter}



\title{Multi-Task Diffusion Approach For Prediction of Glioma Tumor Progression}

\author{Aghiles Kebaili$^a$}

\author{Romain Modzelewski $^{a,c}$}

\author{Jérôme Lapuyade-Lahorgue $^b$}

\author{Maxime Fontanilles $^c$}

\author{Sébastien Thureau $^{a,c}$}

\author{Su Ruan$^{a,c}$}

\affiliation{organization={AIMS Lab, Quantif, University of Rouen-Normandy},
            city={Rouen},
            postcode={76000},
            state={Normandy},
            country={France}}

\affiliation{organization={LITIS Lab UR 4108, University of
Rouen-Normandy},
            city={Rouen},
            postcode={76000},
            state={Normandy},
            country={France}}

\affiliation{organization={CLCC Henri Becquerel},
            city={Rouen},
            postcode={76038},
            state={Normandy},
            country={France}}

\begin{abstract}
Glioma, an aggressive brain malignancy characterized by rapid progression and its poor prognosis, poses significant challenges for accurate evolution prediction. These challenges are exacerbated by sparse, irregularly acquired longitudinal MRI data in clinical practice, where incomplete follow‑up sequences create data imbalances and make reliable modeling difficult. In this paper, we present a multitask diffusion framework for time‑agnostic, pixel‑wise prediction of glioma progression. The model simultaneously generates future FLAIR sequences at any chosen time point and estimates spatial probabilistic tumor evolution maps derived using signed distance fields (SDFs), allowing uncertainty quantification. To capture temporal dynamics of tumor evolution across arbitrary intervals, we integrate a pretrained deformation module that models inter‑scan changes using deformation fields. Regarding the common clinical limitation of data scarcity, we implement a targeted augmentation pipeline that synthesizes complete sequences of three follow‑up scans and imputes missing MRI modalities from available patient studies, improving the stability and accuracy of predictive models. Based on merely two follow‑up scans at earlier timepoints, our framework produces flexible time‑depending probability maps, enabling clinicians to interrogate tumor progression risks at any future temporal milestone. We further introduce a radiotherapy‑weighted focal loss term that leverages radiation dose maps, as these highlight regions of greater clinical importance during model training. The proposed method was trained on a public dataset and evaluated on an internal private dataset, achieving promising results in both cases.



\end{abstract}



\begin{keyword}
Diffusion Models \sep Longitudinal Follow-up \sep Multimodal MRI \sep Tumor Progression Prediction \sep Uncertainty 



\end{keyword}

\end{frontmatter}



\section{Introduction}
\label{sec:introduction}
Gliomas are among the most aggressive and deadliest forms of brain tumors. Gliomas grows rapidly and strongly affects surrounding brain tissue, often causing a "mass effect" that distorts adjacent structures \cite{tuncc2021modeling}. In 2022, the incidence of brain and central nervous system tumors in Western Europe was estimated at 5.56 per 100,000 individuals. This reflects some of the highest rates observed in recent years .
Even with current standard treatments, which typically involves maximal safe resection followed by radiotherapy and temozolomide chemotherapy, patients usually survive only 12 to 15 months, with fewer than 20\% living beyond 5 years according the EUROCARE-5 study \cite{stupp2005radiotherapy}. This poor outcome is mainly due to the tumor’s intrinsic aggressiveness, a high degree of intra-tumoral heterogeneity and diffuse infiltration. Unlike many solid tumors that remain spatially confined, gliomas tend to spread widely. This extensive spread presents a critical challenge in treatment planning, as most conventional therapies typically target the main tumor mass and may miss the scattered and residual infiltrative cells \cite{chicoine1995assessment}. 




Magnetic resonance imaging (MRI) is the most widely used modality for brain tumors progression follow-up due to its high spatial resolution and superior multi-contrast soft tissue representations that allows detailed visualization of tumor boundaries and its associated different tissues \cite{zhou2023prediction,kebaili2025multi}. 


Deep learning methods show strong potential for predicting clinical outcomes across diverse cancer subtypes \cite{cai2024automated}. However, current approaches for modeling brain tumor progression at a pixel level remain insufficiently reliable for clinical use, with persistent accuracy variations across patients in long-term predictions. This stems from two key challenges: the complexity of heterogeneous tumor growth patterns 
and limited availability of longitudinal neuroimaging datasets required to train robust models, contributing to the field’s slow progress. Emerging generative models could address these limitations by synthesizing realistic images to augment predictive models in data-scarce regimes \cite{kebaili2023deep}. Such predictive models may directly improve personalized treatment strategies by quantifying patient-specific progression risks and spatial growth dynamics. Currently, only few studies have been dedicated to this field in the literature.

In this work, we propose a multitask diffusion framework for
flexible, time-agnostic prediction of glioma progression. Section 2 provides a review of related works, while Section 3 details our proposed methodology. Subsequent sections discuss experimental results.

\section{Background}
\label{sec:background}
Over the past few decades, predicting brain tumor progression has primarily relied on three methodologies: mathematical modeling with partial differential equations (PDEs), radiomic feature extraction and analysis, and biomechanical models that simulate tissue deformations. Mathematical models have long been key to understanding the invasive behavior of gliomas. Early works by Tracqui et al. (1995) \cite{tracqui1995passive} and Woodward et al. (1996) \cite{woodward1996mathematical}
introduced frameworks which identify two principal characteristics of tumor growth: cell proliferation and diffusion. In these models, the dynamics of tumor evolution are captured by systems of partial differential equations, commonly using reaction-diffusion equations. 
Over the years, these models have been improved to consider spatial heterogeneity of the tumor \cite{proietto2023tumor}.

As an alternative to reaction-diffusion models, radiomic approaches have emerged as a strong data-driven alternative. Radiomics are defined as quantitative features extracted from medical images. 
These features are then correlated with clinical outcomes using machine learning classifiers to predict how tumors might progress. Early work by Lambin et al. (2012) \cite{lambin2012radiomics} demonstrated that such quantitative imaging biomarkers could improve tumor characterization and growth prediction.


In addition to reaction-diffusion models, other approaches integrate mathematical models of tumor growth with elasticity theories to capture the structural deformations caused by expanding tumors
\cite{subramanian2019simulation,clatz2005realistic}. More recent studies have advanced these models by incorporating nonlinear elasticity frameworks to better account for complex, heterogeneous mechanical properties of brain tissue \cite{wong2015tumor}.


Recent advances have explored deep learning approaches for predicting glioma progression, sometimes in combination with traditional mathematical models. One such approach incorporates a proliferation-invasion framework to generate dose matrices around the tumor, which are then used to guide machine learning predictions of tumor evolution \cite{gaw2019integration}. More recent strategies leverage deep segmentation features to capture tumor morphology and spatial distribution. For example, Elazab et al. introduced GP-GAN, a stacked 3D generative adversarial network conditioned on segmentation features to synthesize future tumor masks from MR images, although its development on a small cohort of only 18 patients limits its clinical applicability \cite{elazab2020gp}. Some approaches further incorporate temporal modeling to account for irregular follow-up intervals \cite{petersen2021continuous,petersen2019deep}. A notable example is the treatment-aware diffusion probabilistic model, which integrates treatment data and time-conditioned segmentation to forecast tumor evolution. Despite its innovation, this model was trained on a private dataset with very short intervals between follow-up scans, yielding only subtle tumor changes that make temporal interpolation overly smooth, and on long longitudinal scan sequences that are rarely available in standard clinical workflows \cite{liu2025treatment}. Despite their promise, these deep learning methods face several challenges that hinder clinical deployment. They require large, high-quality longitudinal datasets, which are often unavailable due to inconsistent imaging protocols and sparse follow-up scans. The heterogeneity in tumor behavior and treatment response further limits model generalization.

\subsection*{Contributions}

This work introduces a multitask diffusion framework for flexible, time‑agnostic prediction of glioma progression. Given just two prior scans, the model can generate future FLAIR MRIs at any chosen timepoint and produce pixel‑level probability maps of tumor evolution. We propose to use signed distance maps, computed from the initial tumor contour, to represent spatial uncertainty. To accommodate uneven changes in tumor shape between scans, we integrate a pretrained tumor evolution module that captures spatial tumor changes as a dense vector field. We also address the scarcity of longitudinal data through a targeted augmentation pipeline, synthesizing entire sequences of three follow‑ups and imputing missing MR sequences from existing ones. Finally, a radiotherapy‑weighted focal loss guides the model to prioritize clinically critical regions during optimization. Our contributions are:

\begin{itemize}
    \item Multi-task model to predict glioma progression at arbitrary future timepoints, with uncertainty modeling.
    \item Tumor evolution module to model spatial tumor changes and evolution across different scan intervals.
    \item A data augmentation pipeline to overcome limited longitudinal data and missing MR sequences.
    \item Radiotherapy‑weighted focal loss by leveraging dose maps to focus learning on high‑risk tumor regions and for a faster convergence.
    \item Promising results are obtained on the public dataset, with strong generalization demonstrated on an independent private test set.
\end{itemize}

\section{Preliminary}
\label{sec:preliminary}
Diffusion models \cite{ho2020denoising} are a subset of generative models based on a forward-and-backward diffusion process. 
During the forward, Gaussian noise is gradually added to an initial data point $x_0 \sim q(x_0)$ from the target data distribution, following a predefined variance scheduler $\beta_1, \dots, \beta_T \in [0, 1]$:

\begin{equation}
    q(x_t|x_{t-1}) := \mathcal{N}(x_t; \sqrt{1 - \beta_t} x_{t-1}, \beta_t \mathbf{I}),
\end{equation}


where $t \in \{1, ..., T\}$ denotes the timestep, $\beta_t$ is the variance at timestep $t$, and $\mathbf{I}$ is the identity matrix. 
During backward, we approximate the reverse diffusion process $q(x_{t-1}|x_t)$, which allows us to gradually transform Gaussian noise $x_T \sim \mathcal{N}(\mathbf{0}, \mathbf{I})$ into the initial data distribution $q(x_0)$. However, since $q(x_{t-1}|x_t)$ is intractable, a model with paramters $\theta$ is introduced to approximate the Gaussian mean $\mu_\theta(x_t, t)$ at each timestep $t$. This process can be expressed as:

\begin{equation}
    p_\theta(x_{t-1}|x_t) = \mathcal{N}(x_{t-1}; \mu_\theta(x_t, t); \sigma^2_t )
\end{equation}


The model learns the parameters $\theta$ by maximizing the evidence lower bound for each of the backward Markov chain steps, similarly to VAEs \cite{kingma2013auto}. As described in \cite{ho2020denoising}, the loss function is further simplified into $L_\text{simple}$ defined as follow: 

\begin{equation}
     \mathcal{L}_{simple} := \mathbb{E}_{t, \epsilon} || \epsilon - \epsilon_\theta(x_t, t) ||^2
    \label{eq:loss}
\end{equation}
New data points $\tilde{x} \sim q(x_0)$ can be sampled by diffusing a random noise vector $x_T \sim \mathcal{N}(\mathbf{0}, \mathbf{I})$ through all steps of the Markov chain (hundereds to thousands). 

\section{Method}
\label{sec:method}
Our approach operates within a multi-task diffusion-based framework that extracts core features from the inputs, shared then between a generative branch and a predictive branch. We first propose an architecture named GliomaDiff, which formulates glioma evolution as a conditional prediction task: given past scans $\{S_{t_1}, S_{t_2}\}$ and their acquisition days $\{d_{t_1}, d_{t_2}\}$, it predicts a future scan $S_{t_3}$ and the corresponding probability map $\hat{P}_{t_3}$ of tumor evolution at a given arbitrary timepoint $d_{t_3}$. As its core, GliomaDiff relies on two key probabilistic representations. First, hard-labeled tumor masks are encoded as logistic-normalized signed distance fields, which provide soft-labeled maps, enabling uncertainty quantification. Second, the tumor evolution representation uses predicted deformation fields to capture nonlinear spatial changes of the tumor between timepoints. Given the typical scarcity of longitudinal data, we integrate a data augmentation pipeline based on synthetic images, enriching the real training samples with additional plausible tumor evolution trajectories.

\subsection{GTV Probabilistic Representation via Signed Distance Fields}

Our main goal is to predict future tumor locations from earlier scans. This task fundamentally differs from traditional segmentation problems that rely on static, per-pixel classification of a source image. Segmentation typically treats each pixel as a hard label (tumor or no tumor) given the source image, while we must infer the mask purely from prior follow-ups. Consequently, binary representations can be limiting when the target is not a direct mapping segmentation from a known source image, but rather a plausible future state of the tumor \cite{lourencco2021using}. Predicting a precise binary tumor contour introduces rigid constraints that may limit our predictive capabilities, as growing tumors are variable and uncertain. Clinically, it is also questionable to interpret these hard predictions as absolute, given the inherent uncertainty in future tumor behavior. Therefore, instead of enforcing the prediction of exact binary boundaries, we adopt a probabilistic representation of the Gross Tumor Volume (GTV), which better reflects the uncertainty and gradual spatial spread of tumor presence. To give the model more flexibility and to reflect uncertainty near tumor edges, we reformulate mask prediction as probabilistic regression. In this setup, the network outputs a probability map $P(x)$, estimating at each pixel the likelihood of tumor presence given earlier scans. This approach lightens binary contour constraints and lets the model learn a more flexible spatial distribution of tumor evolution particularly under limited data, case where uncertainty is highly present.

To achieve this, we convert each binary GTV mask into a continuous-valued probability map $P_t$ using signed distance functions (SDFs). At each pixel $x$, $P_t(x)$ represent the likelihood of the tumor presence. Let the GTV mask be defined as $M_t: \Omega \rightarrow \{0, 1\}$, where $\Omega$ is the pixel space, and $t$ is the current follow-up timepoint. We compute the SDF as follow :

\begin{equation}\label{eq:sdf}
    \phi(x) = \begin{cases}
        -\text{min}_{y\in \partial \mathcal{T}} ||x - y||, \;\;\textbf{if}\;\; x \in \mathcal{T}\\
        +\text{min}_{y\in \partial \mathcal{T}} ||x - y||, \;\;\textbf{if}\;\; x \notin \mathcal{T}
    \end{cases}
\end{equation}

Where $\mathcal{T} = \{x \in \Omega | M_t(x) = 1\}$ is the set of tumoral pixels, and $\partial \mathcal{T}$ is the boundery of the tumor. The further a pixel is from the boundery, the less likely it is to belong to the tumor. We propose to use this signed distance driven by a smooth logistic decay function to model the tumor probability at each location. Indeed, Logistic functions have a long history in modeling tumor growth and invasion dynamics, providing a biologically plausible decay profile away from the core\cite{yankeelov2013clinically,jarrett2018mathematical}. To this end, we define the probability function for a given pixel $x$ belonging to the tumor as  :

\begin{equation}
    \label{eq:logistic}
    P_t(x) = \sigma(\phi(x), \beta, \mu) = \frac{1}{1 + e^{\beta(x - \mu)}}
\end{equation}

\begin{figure}[t]
    \centering
    \includegraphics[width=\columnwidth]{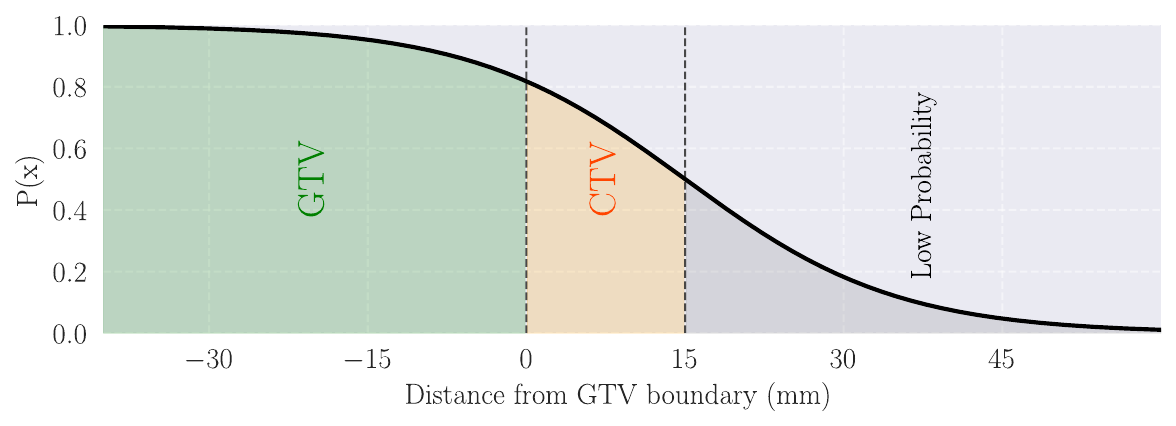}
    \caption{Logistic normalization function $P_t(x)$ w.r to the distance from the GTV boundary SFD map $\phi$, using a $\beta=0.1$ and $\mu=15$. Clinical regions highlighted: GTV ($d < 0mm$) distance from tumor boundaries is in green, CTV ($0mm \leq d \leq 15mm$) in orange, and low-probability region ($d > 15mm$) in gray.}
    \label{fig:sigmoid}
\end{figure}

The parameter $\beta$ controls how sharply the probability drops off, while $\mu$ centers the curve such as $P_t(\mu) = 0.5$. Based on empirical observations from our dataset and clinical practice, we configure this function so that the tumor core maintains a high probability (i. e, $P_t > 0.8$), with the probability decaying toward 0.5 at approximately 15 mm from the boundary ($\beta = 0.1, \mu = 15$). This $15mm$ margin aligns with clinical guidelines \cite{baumert2025estro}, where the Clinical Target Volume (CTV) is often defined as $\text{CTV} = \text{GTV} + 15mm$ of safety margin to take into consideration the microscopic spread \cite{di2024impact}. Below $P_t = 0.5$, we consider tumor presence unlikely. This representation provides GliomaDiff with a softer, more informative learning signal, allowing it to capture the gradual transition between tumor and healthy tissue, while also offering greater flexibility and facilitating uncertainty modeling (see Figure \ref{fig:sigmoid}).

\begin{figure*}[t]
    \centering
    \includegraphics[width=\textwidth]{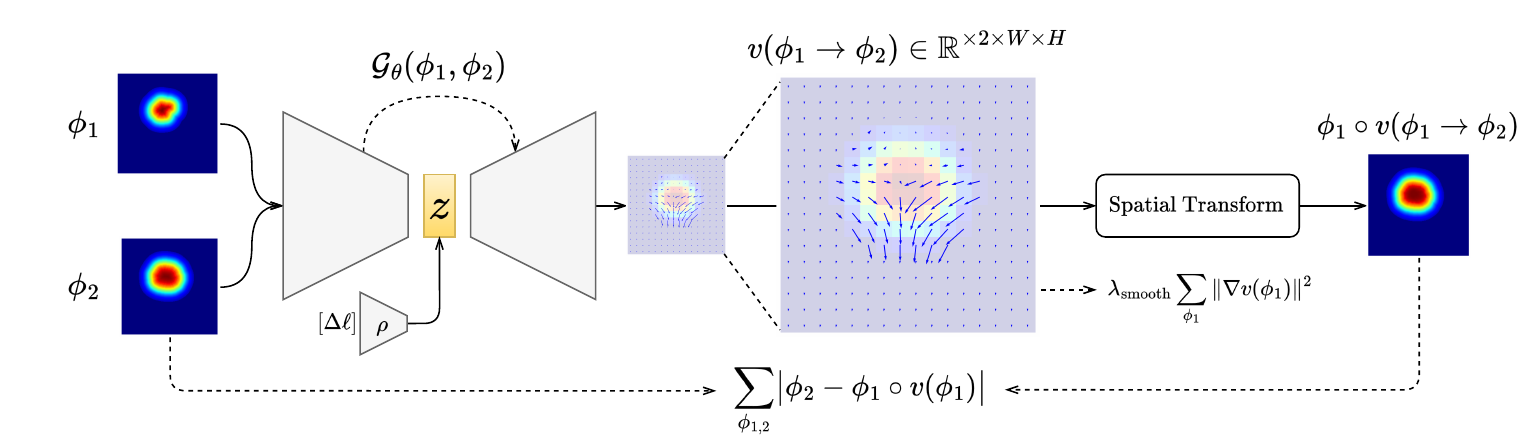}
    \caption{Overview of our tumor evolution module. Given two distance maps $\phi_1$ and $\phi_2$ and a time gap $\Delta\ell$ encoded via the $\rho$-embedding, our U-Net–style network $\mathcal{G}_\theta$ predicts a dense vector field $v(\phi_1 \rightarrow \phi_2)$, describing the spatial tumor evolution. This field is applied to $\phi_1$ through a spatial transformer to align it with $\phi_2$.}
    \label{fig:architecture_deformation}
\end{figure*}

\subsection{Tumor Evolution Module with Learned Deformation Fields}
Accurately capturing how a tumor changes between two timepoints is essential for a reliable progression prediction. Recurrent multi-scan networks implicitly model temporal evolution by processing sequences of scans, but they typically need more than two images and are prone to overfitting when data are scarce. In contrast, our framework uses a single-source to single-target deformation architecture, given the limited number of follow-up scans available. Our deformation network $\mathcal{G}_\theta$ inspired from \cite{balakrishnan2019voxelmorph}, is defined as a U-Net and takes the two SDF maps $\phi_1$ (at time $t_1$) and $\phi_2$ (at time $t_2$) as input and predicts a dense 2D deformation field $v(.)$, that warps the first map into the second. To focus exclusively on tumor evolution and avoid noisy deformations of healthy brain tissue, the network processes only the SDF (pre-logistic normalization) for each timepoint rather than full MR images. We also supply the model with temporal information by encoding the time gap $\Delta \ell = t_2 - t_1$ in days, enabling the network to scale its output for arbitrary intervals and improve its generalizability. The deformation model outputs for each pixel $x, y$, a 2D displacement vector $v_{x,y} \in \mathbb{R}^2$. We can apply a spatial transform (defined with the $\circ$ operator) to obtain $\phi_2 = \phi_1\circ v(\phi_1 \rightarrow \phi_2)$ during training, which remains fully differentiable.

Training aligns the warped map $\phi_1 \circ v(\phi_1 \rightarrow \phi_2)$ with the target map $\phi_2$ by minimizing a similarity loss, while enforcing a smooth, physicall plausible deformation by penalizing large gradients in the displacement field. The loss function is defined as follow:

\begin{equation}
\mathcal{L}_{\mathrm{def}}(\phi_1, \phi_2) = \sum_{\phi_{1,2}} \bigl|\phi_2 - \phi_1 \circ v(\phi_1)\bigr| \quad + \lambda_{\mathrm{smooth}} \sum_{\phi_1} \|\nabla v(\phi_1)\|^2
\end{equation}

Figure \ref{fig:architecture_deformation} depicts the detailed architecture for the tumor evolution module, which at inference time generates $v(\phi_1 \rightarrow \phi_2)$ on the fly for any two scans and provides a deformation prior for our downstream prediction model.

\subsection{Multi-Task Diffusion Prediction Model}
We define our prediction network as a two-branch multi-task diffusion model operating in pixel space, designed to simultaneously generate future FLAIR scans and predict tumor evolution as pixel-wise probability maps. The model uses a shared U-Net encoder with two separate decoder heads:
\begin{itemize}
    \item \textbf{Generative branch}: Trained to synthesize the target-time FLAIR image (limited to FLAIR by dataset constraints; see Section \ref{sec:experiments:datasets}).
    \item \textbf{Predictive branch}: trained to output a continuous tumor probability map for the same timepoint.
\end{itemize}

Both decoders share skip connections from the encoder to maximize feature reuse. In addition, the network is conditioned by the deformation field predicted by our tumor evolution module to guide spatial transformations. 

\textbf{Temporal Conditioning:} To incorporate temporal context, we encode the relative scan intervals $\Delta \ell_1$ (between $t_1$ and $t_2$) and $\Delta \ell_2$ (between $t_2$ and the target timepoint) via a sinusoidal positional embedding. These embeddings are injected throughout the network using FiLM layers \cite{perez2018film}, modulating both encoder and decoder activations so that the model adapts its outputs according to the specific time gaps.

\begin{figure*}[t]
    \centering
    \includegraphics[width=\textwidth]{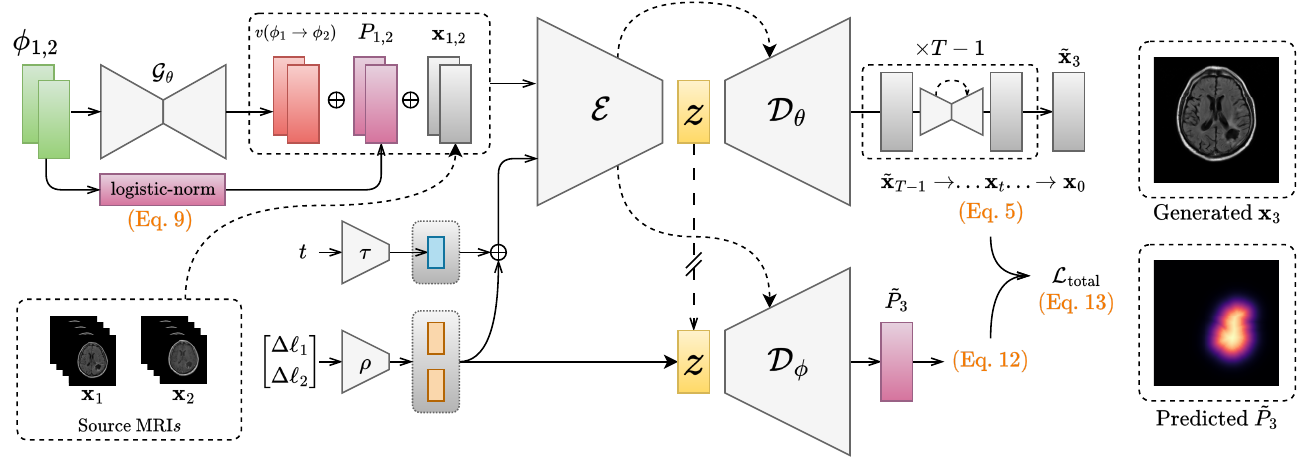}
    \caption{Detailed illustration of our prediction architecture. Source inputs are defined as real+synthetic \textbf{FLAIR}, \textbf{T1CE}, imputed \textbf{T1, T2} and $\mathbf{\phi}$ for each of $t_1$ and $t_2$ sequences, alongside the \textbf{RTDOSE} map. Diffusion timestep $t$ and time condition $\Delta\ell$ are encoded through $\tau$ and $\rho$-embeddings. The prediction branch takes only the $\Delta\ell$ embedding. }
    \label{fig:architecture_globale}
\end{figure*}

The two decoder branches are trained jointly under a single composite loss. The generative branch is supervised by the standard diffusion reconstruction loss, $\mathcal{L}_\text{diff}$, which measures the denoising error at each diffusion timestep:

\begin{equation}
    \mathcal{L}_\text{diff} = \mathbb{E}_{x_t, \epsilon, \Delta \ell, v(x)} \left[ ||\epsilon - \epsilon_\theta(x_t, t, \Delta \ell, v(x))||^2 \right]
\end{equation}

The predictive branch is based on a radiotherapy-weighted term $\mathcal{L}_\text{RT}$, similar to the focal loss \cite{lin2017focal}, which emphasizes high-risk tumor regions during training given the radiation dose map:
\begin{equation}
    \mathcal{L}_\text{RT} = \left(1 + R^\gamma\right) \cdot \bigl[P(x)\,\log\bigl(\hat{P}(x)\bigr) + \bigl(1 - P(x)\bigr)\,\log\bigl(1 - \hat{P}(x)\bigr)\bigr]
\end{equation}

where $\gamma > 0$ is the focal exponent controlling the sensitivity to high-dose regions. The overall loss function is then defined as:

\begin{equation}
    \mathcal{L}_{\mathrm{total}} = \mathcal{L}_{\mathrm{diff}} + \lambda_{\mathrm{RT}} \cdot \mathcal{L}_{\mathrm{RT}}
\end{equation}

where $\lambda_{\mathrm{RT}}$ is a hyperparameter that balances image fidelity against targeted prediction accuracy, fixed at 0.1 in our experiments. Figure \ref{fig:architecture_globale} illustrates the full multi-task diffusion architecture, including how the encoder feeds both decoders and how each loss term is applied to its respective branch.

\subsection{Data Augmentation Pipeline}
To overcome the scarcity of longitudinal MRI data, we integrate a two-stage, diffusion-based augmentation pipeline into our prediction framework:

\textbf{Full-Sequence Synthesis}: We leverage a pixel-space denoising diffusion model \cite{ho2020denoising,kebaili2025multi} to generate entire three-scan sequences along with their corresponding GTV probability maps and radiation dose maps. This model produces high-quality synthetic follow-ups that expand the diversity of tumor evolution trajectories available for training. The diffusion model is also conditioned on the temporal gap between scans $\Delta \ell$, allowing it to learn the underlying correlation between the tumor growth and temporal information.

\textbf{Modality Imputation}: A second diffusion network performs missing-sequence imputation inspired from our previous works \cite{kebaili2025amm}, taking any available scans as input and synthesizing the absent MRI modalities. This ensures that, even when patient records lack certain timepoints or sequences, the prediction model receives a complete, temporally coherent set of inputs.

Note that our Full‐Sequence generator is trained exclusively on the limited real follow‐up data available, that is, it synthesizes only three‐timepoint FLAIR and T1CE sequences, as those sequences are the only available ones (see Section \ref{sec:experiments:datasets}). Our pretrained imputation model then fills in the missing T1 and T2 modalities. This two‐stage strategy avoids training the generative network on purely synthetic data— to avoid introducing a distribution shifts, which can cause the model to collapse \cite{shumailov2024ai}. We also carefully calibrate the ratio of real to synthetic examples during training to prevent over‐reliance on synthetic images \cite{kebaili2023deep}. Further discussion of the augmentation pipeline and its effect on performance is provided in Section \ref{sec:results}.

\section{Experiments}
\label{sec:experiments}

\subsection{Datasets}
\label{sec:experiments:datasets}
We base our study on the publicly available Burdenko Glioblastoma Progression Dataset (BGPD), which includes imaging and clinical data from 180 patients treated for primary glioblastoma between 2014 and 2020 at the Burdenko National Medical Research Center of Neurosurgery \cite{Zolotova2023-ye}. The dataset contains a total of 645 imaging studies, with a minimal set of T1CE and FLAIR sequences per follow-up. These studies are highly heterogeneous in structure, with the number of follow-up scans per patient ranging from 1 to 8, with an average of 4 scans, including the radiotherapy planning timepoint. To maximize the number of usable patients and standardize the input format, we limit our analysis to 3 or more timepoints per patient. Extending the minimal sequence length to 4 timepoints would shrink the cohort to only 49 patients, which is too small for effective training. After excluding aberrant data, we end up with a cohort of 112 patients.


\textbf{Preprocessing:} Gross Tumor Volumes (GTVs) are mostly delineated on FLAIR sequences. Nearly half of the selected follow-up studies lack GTV annotations. To address this, we trained a 3D nnU-Net segmentation model \cite{isensee2018nnu} on the available labeled data using FLAIR and T1CE as input modalities. This model achieved a mean Dice score of 0.76 on a held-out test set and was used to complete the missing GTVs. The majority of MRI scans in BGPD are anisotropic, with slice thickness reaching up to 5 mm and volumes containing as few as 20 axial slices. The remaining MRIs were rigidly registered to their corresponding CT scans, then resized to a fixed shape of $256\times256$. Each final study in our dataset includes FLAIR and T1CE sequences, RTDOSE maps, and either manual or automatically generated GTV masks. 


\textbf{Local data for evaluation:} 
To complement the BGPD cohort, we assembled a local evaluation set of five patients treated at our cancer center between 2016 and 2020. For each patient, we retrieved three longitudinal studies, one radiotherapy planning scan and two follow-up timepoints: each comprising T1CE and FLAIR MR images, co-registered to the planning CT and accompanied by the corresponding RTDOSE map. GTVs were manually delineated on the planning scans; for the follow-up studies, we applied the same pretrained 3D nnU-Net model used on BGPD to automatically segment the GTVs.

\subsection{Evaluation Metrics}
To assess the accuracy of our tumor progression predictions, we use a suite of complementary metrics. For the prediction task, we use: 1) Root mean square error (RMSE) and binary cross-entropy (BCE) between the ground-truth and predicted probability maps to evaluate pixel-wise deviations. 2) Kullback–Leibler (KL) divergence to measure how much the predicted probabilistic GTV distribution differs from the ground truth. 3) Expected Calibration Error (ECE) to quantify how well predicted confidences match observed frequencies across probability bins, capturing the model’s uncertainty calibration. 4) Dice score (DSC), computed after thresholding the predicted maps at $P_t > 0.8$, to assess spatial overlap between predicted and true tumor regions. For image synthesis quality, we use both pixel-level and perceptual metrics. Peak signal-to-noise ratio (PSNR) and structural similarity index (SSIM) evaluate how closely the generated FLAIR scans match the originals in terms of pixel accuracy and structural integrity. For perceptual quality, we include LPIPS which compares deep features to estimate visual similarity.

\subsection{Implementation Details}
\label{sec:imlplementation_details}
The downstream prediction model is a U-Net–style network with four downsampling and upsampling stages (factor $f = 4$) and residual convolutions with a $3\times3$ kernel size and $64, 128, 256, 512$ filters at each stage. The model architecture consists of one shared encoder, and two parallel decoders: A generative and predictive branch (Figure \ref{fig:architecture_globale}), conditioned on the scan gaps $[\Delta \ell_1, \Delta \ell_2]$ via a sinusoidal positional embedding followed by an MLP that modulates the network's activations with FiLM layers \cite{perez2018film}. Our data augmentation pipeline also uses two diffusion networks from our prior work \cite{kebaili2025multi,10980985}: The AMM-Diff for missing modality imputation and a pixel-space latent diffusion model for synthesizing complete three-timepoint FLAIR, T1CE, RTDOSE map, and GTV sequences. The tumor evolution module is also a U-Net-like architecture with a four-stage down/upsampling blocks, conditioned on the time gaps (Figure \ref{fig:architecture_deformation}). During its training, we randomly sample pairs of scans from the same patient while additionally feed in their pre-logistic signed‐distance GTV maps (Eq. \ref{eq:sdf}) to preserve smooth, continuous deformation vectors.

We train GliomaDiff using three scans per patient: two consecutive scans as inputs and a randomly selected futur scan as target. This flexible sampling strategy yields 866 unique triplets improving the coverage of tumor evolutions. To prevent condition memorization, we add $\pm 5$ day noise to the $[\Delta \ell_1, \Delta \ell_2]$ conditioning variables. We exclude low-information slices with $<75 mm^2$ tumor area to keep only clinically relevant examples. The sequence synthesis network trains on BGPD using three-scan sequences (following our earlier sampling scheme), with each scan containing FLAIR, T1CE, probabilistic GTV maps, and patient-specific RTDOSE maps, conditioned on temporal vector $[\Delta \ell_1, \Delta \ell_2]$. Our AMM-Diff imputation model leverages transfer learning and is pretrained on the public BraTS dataset to recover missing T1 and T2 modalities. We split our 112 patient cohort into 105 for training and 7 for testing. All experiments were conducted on 8 NVIDIA A100 80 GB GPUs, for 48 hours using a batch size of 16. The prediction model is trained using Adam optimizer with a $4 \times 10^{-5}$ learning rate and a linear scheduler with 1,000 diffusion steps. We use the nnU-Net \cite{isensee2018nnu} framework for the autoGTV segmentations.


\subsection{Results}
\label{sec:results}

\begin{figure}[t]
\centering
    \includegraphics[width=\columnwidth]{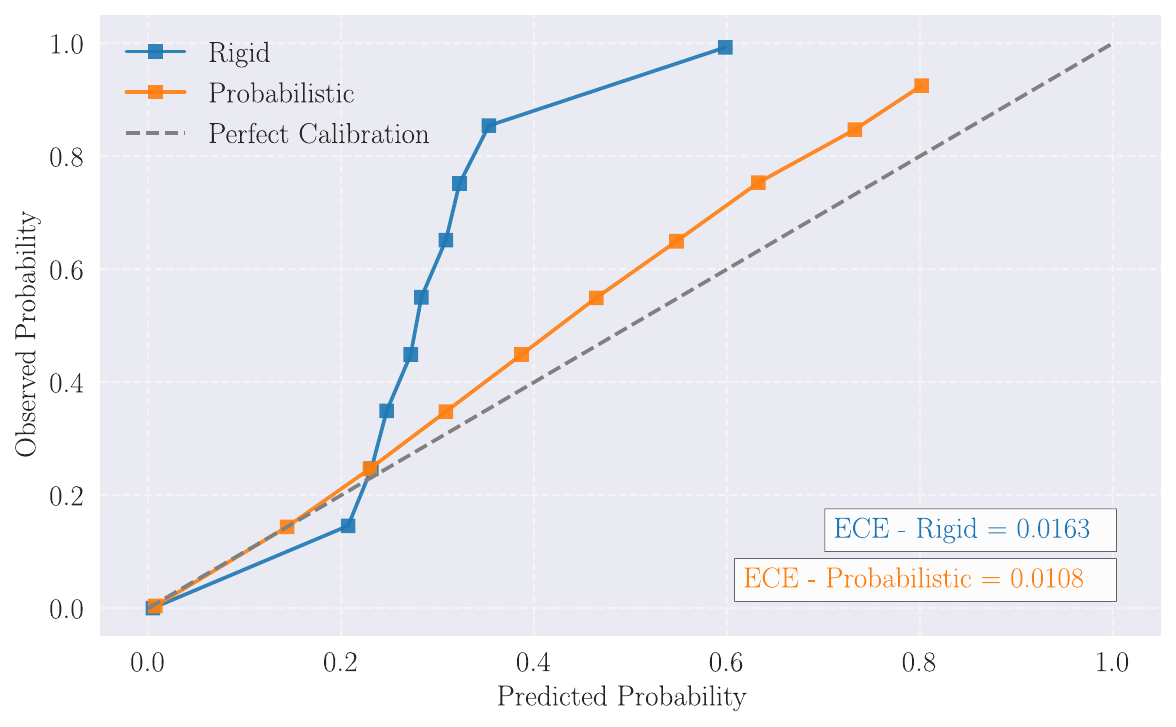}
    \caption{Calibration curves between observed and predicted probabilities, comparing the binary GTV masks versus our logistic-normalized probabilistic maps. The probabilistic approach achieves tighter calibration across all probability bins, reducing the uncertainty.}
    \label{fig:rigid_vs_prob}
\end{figure}

The test cohort consists of 7 patients, selected to reflect the overall follow-up distribution in our BGPD dataset (71\% with 3 scans, 17\% with 4-5, and 12\% with more than 5) constructing 23 unique prediction combinations following our sampling scheme (Section \ref{sec:imlplementation_details}), averaging around 5 tumoral slices per combination, we finally obtain 300 evaluation points. We first evaluate our method between binary and probabilistic predictions, we replace binary GTV masks with our logistic-normalized probabilistic maps and measure the resulting improvement on a standard pixel-space diffusion model. Table \ref{tab:rigid_vs_prob} shows that this probabilistic representation lowers the RMSE, CE and ECE. Hard-label CE has a zero-loss floor for perfect matches, whereas soft-label CE cannot reach zero even under perfect alignment, consequently, these scores sit on slightly different scales. CE serves solely as a rough proxy for prediction quality rather than an absolute performance scale. To provide a direct overlap metric, we report the DSC score by thresholding predictions at $P_t > 0.8$, this cutoff chosen to mirror the logistic transform’s mapping of true tumor pixels above 0.8 (Eq. \ref{eq:logistic}). We see improvements in RMSE, CE and ECE alongside a small gain in DSC, confirming tighter spatial agreement with ground truth. Figure \ref{fig:rigid_vs_prob} further illustrates the reliability diagram and calibration curves, showing a consistent reduction in ECE across all probability thresholds, droping from 0.0163 with binary GTVs to 0.0100 (improvement of 38\%) with probabilistic maps. This demonstrates our framework’s superior ability to continuously quantify uncertainty.

\begin{table}[h]
\caption{Comarison of performance with binary GTV masks versus logistic-normalized probabilistic GTV maps on the test cohort}
\label{tab:rigid_vs_prob}
\centering
\resizebox{\columnwidth}{!}{%
\small
\begin{tabular}{@{}lcccc@{}}
\toprule
Methods & RMSE $\downarrow$ & CE $\downarrow$ & DICE $\uparrow$ & ECE $\downarrow$\\ \cmidrule(r){1-1} \cmidrule(lr){2-2} \cmidrule(lr){3-3} \cmidrule(lr){4-4} \cmidrule(l){5-5}
Binary & 0.111 $\pm$ 0.02 & 1.387 $\pm$ 0.80 & 0.683 $\pm$ 0.13 & $0.016$ \\
Probabilistic & \textbf{0.062 $\pm$ 0.03} & \textbf{0.108 $\pm$ 0.04} & \textbf{0.689 $\pm$ 0.12} & $\mathbf{0.010}$\\
\bottomrule
\end{tabular}
}
\end{table}

\begin{table*}[t]
\caption{Ablation performance of the different modules on predictive and generative performance. Second best are underlined.}
\label{tab:ablation}
\centering
\resizebox{\textwidth}{!}{%
\small
\begin{tabular}{@{}lcccccccc@{}}
\toprule
\multirow{2}{*}{Method} & \multicolumn{4}{c}{Prediction metrics} & \multicolumn{4}{c}{Generative metrics} \\
\cmidrule(lr){2-5} \cmidrule(l){6-9}
 & RMSE $\downarrow$ & CE $\downarrow$ & KL $\downarrow$ & DICE $\uparrow$ & MSE $\times 10^{-3}$ $\downarrow$ & PSNR $\uparrow$ & SSIM $\uparrow$ & LPIPS $\downarrow$ \\
\cmidrule(r){1-1} \cmidrule(lr){2-2} \cmidrule(lr){3-3} \cmidrule(lr){4-4} \cmidrule(lr){5-5} \cmidrule(lr){6-6} \cmidrule(lr){7-7} \cmidrule(lr){8-8} \cmidrule(l){9-9}
Probabilistic GTV (Baseline) & 0.062 $\pm$ 0.03 & 0.108 $\pm$ 0.04 & 0.540 $\pm$ 0.53 & 0.689 $\pm$ 0.12 & $0.391 \pm 0.21$ & $24.780\pm 2.72$ & $0.747 \pm 0.07$ & $0.073 \pm 0.01$\\
\cmidrule(r){1-1} \cmidrule(lr){2-2} \cmidrule(lr){3-3} \cmidrule(lr){4-4} \cmidrule(lr){5-5} \cmidrule(lr){6-6} \cmidrule(lr){7-7} \cmidrule(lr){8-8} \cmidrule(l){9-9}
+ RT-Weighted Loss & 0.056 $\pm$ 0.03 & 0.137 $\pm$ 0.08 & 0.381 $\pm$ 0.38 & 0.685 $\pm$ 0.13 & $\underline{0.374 \pm 0.18}$ & $25.200 \pm 2.44$ & $0.758 \pm 0.06$ & $0.071 \pm 0.01$\\
+ RT-Weighted + Deformation & 0.055 $\pm$ 0.03 & 0.128 $\pm$ 0.05 & 0.336 $\pm$ 0.38 & 0.694 $\pm$ 0.10 & $\mathbf{0.355\pm 0.19}$ & $\underline{25.642\pm 2.84}$ & $\underline{0.804 \pm 0.08}$ & $\mathbf{0.068 \pm 0.01}$\\
+ RT-Weighted + Def + Std. Aug & $\underline{0.051 \pm 0.03}$ & \textbf{0.098 $\pm$ 0.03} & 0.272 $\pm$ 0.23 & \underline{0.714 $\pm$ 0.09} & $0.459\pm 0.35$ & $24.895\pm 3.57$ & $0.789\pm 0.09$ & $0.074\pm 0.01$\\
\cmidrule(r){1-1} \cmidrule(lr){2-2} \cmidrule(lr){3-3} \cmidrule(lr){4-4} \cmidrule(lr){5-5} \cmidrule(lr){6-6} \cmidrule(lr){7-7} \cmidrule(lr){8-8} \cmidrule(l){9-9}

+ RT-Weighted + Def + 25 Synth & \textbf{0.048 $\pm$ 0.02} & 0.114 $\pm$ 0.04 & \textbf{0.204 $\pm$ 0.16} & \textbf{0.736 $\pm$ 0.07} & $0.385 \pm 0.24$ & $\mathbf{25.710 \pm 2.52}$ & $\mathbf{0.809 \pm 0.07}$ & $\underline{0.070 \pm 0.01}$\\
+ RT-Weighted + Def + 50 Synth & 0.052 $\pm$ 0.02 & 0.114 $\pm$ 0.04 & \underline{0.250 $\pm$ 0.17} & 0.712 $\pm$ 0.07 & $0.409 \pm 0.24$ & $24.934 \pm 2.52$ & $0.782 \pm 0.07$ & $0.070 \pm 0.01$\\
+ RT-Weighted + Def + 100 Synth & 0.053 $\pm$ 0.02 & \underline{0.104 $\pm$ 0.03} & 0.269 $\pm$ 0.18 & 0.696 $\pm$ 0.09 & $0.447 \pm 0.24$ & $25.265 \pm 2.52$ & $0.774 \pm 0.07$ & $0.073 \pm 0.01$ \\
\bottomrule
\end{tabular}
}
\end{table*}

Table \ref{tab:ablation} presents an ablation study detailing the incremental contributions of each component in our proposed method, with a dedicated section isolating the impact of our synthetic data augmentation for the downstream prediction task. We observe a stable and progressive improvement in prediction performance as we iteratively build upon the baseline model. Overall, each component contributes to progressively improve prediction accuracy. The introduction of the RT-weighted loss term, designed to act as a focal mechanism, modulates the pixel-wise loss by emphasizing high-probability tumor regions while down-weighting background areas, simplifying the learning of low-density structures. Quantitatively, this leads to a noticeable drop in RMSE (0.056) and KL (0.381), indicating better alignment in the overall predictions, though CE slightly increases to 0.137 and DSC score shows a small dip to 0.685. While the gains are not uniformly significant across all metrics, the KL improvement suggests a more coherent global distribution of predicted probabilities. The impact of the RT-weighted term extends beyond prediction metrics, as it positively influences all generative metrics and facilitates faster convergence toward sharper, cleaner generations, likely due to its ability to suppress background noise and focusing on the cranial region, with an emphasis on tumoral structures.


Adding our tumor evolution module further reduces RMSE to 0.055, CE to 0.128, and KL divergence to 0.336, while slightly boosting DSC to 0.694 over the baseline. This demonstrates that modeling tumor evolution as a dense vector field provides a richer representation of tumor dynamics and better captures inter-scan correlations between timepoints. 
On the other hand, generative metrics do not strictly follow the same progressive improvement pattern as the prediction metrics, since the added modules only benefits the prediction branch. The generative branch behaves like a typical diffusion model and acts as a secondary task to help improve the tumor evolution predictions. The overall generative quality nonetheless does slightly improve as each module builds upon the last.

Finally, in the last section of Table \ref{tab:ablation}, we assess the impact of augmenting our training set with synthetic sequences. We vary the number of synthetic scans in $\{25, 50, 100\}$ and find that adding 25 delivers the best overall performance: RMSE $=0.048$, KL $=0.204$, DSC $=0.736 (\uparrow 6.8\%)$, and the highest image-quality scores (PSNR $=25.71$, SSIM $=0.809$, LPIPS $=0.070$). Adding more than 25 sequences negatively impacts to worse both predictive and generative results, showing that simply increasing synthetic data is not always beneficial. This tendency, also reported in our prior work \cite{kebaili2024discriminative}, arises from an imbalance between real and generated scans: synthetic examples, while helpful, remain distinguishable from real ones, and an excess of synthetic over real data shifts the training distribution, causing the model to overfit and degrade overall performance. Generative results confirm this effect, with PSNR and SSIM falling as more generated data are added. These findings underscore a key limitation of synthetic-data augmentation: its benefit depends heavily on the volume of real data available. Overall, augmenting with 25 synthetic follow-ups provides the best performance and boosts prediction results to a final RMSE $=0.048 (\uparrow23\%)$ and DSC $=0.736 (\uparrow6.8\%)$, a significant improvement over the baseline. Furthermore, the constant descrease in the standard deviation of the the whole set of metrics, as well as the ECE $=0.0100 (\uparrow38\%)$, indicates a more stable and reliable model performance and higher confidence in the predictions. This is particularly important in clinical settings, where uncertainty quantification is crucial for treatment planning.

\begin{figure*}[t]
    \begin{center}
    \includegraphics[width=\textwidth]{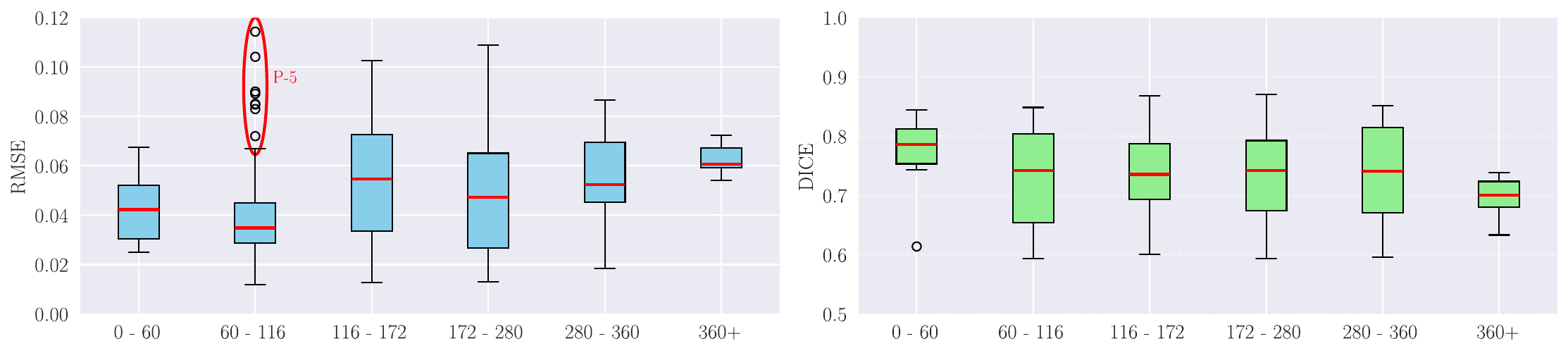}
    \caption{Boxplots of RMSE (left) and DSC (right) across six temporal clusters defined by the interval (in days) between the last input scan and the target. Each box shows the median (red line), interquartile range, and whiskers; open circles denote outliers.}
    \label{fig:delta_t_boxplot}
    \end{center}
\end{figure*}

We next examine how prediction quality varies with the time gap $\Delta \ell_2$ between the last input ($t_2$) and the target ($t_3$). Table \ref{tab:time_intervals} groups the test set into six clusters (0–60, 61–120, 121–180, 181–280, 281–365, $>366$ days), defined by spatially clustering samples according to their $\Delta \ell_2$ values. We observe an inverse correlation between $\Delta \ell_2$ and prediction accuracy: the shortest interval (0–60 days) achieves the lowest RMSE (0.042) and highest DSC (0.781), while the longest interval ($>365$ days) shows the highest RMSE (0.062) and lowest DSC (0.699). RMSE peaks around the 121–180 day window before slightly decreasing at larger $\Delta \ell_2$. Overall, predictions remain reliable across all intervals, though error and uncertainty grow with longer gaps and DSC falls below 0.70 beyond one year. These results indicate that our model forecasts tumor progression accurately up to 12 months ahead. This holds particularly for patients with continuous, predictable growth dynamics, whereas those exhibiting erratic patterns are harder to predict. Figure \ref{fig:delta_t_boxplot} presents boxplots of RMSE and DSC across each temporal interval, illustrating the gradual rise in RMSE and drop in DSC as $\Delta \ell_2$ increases, and highlights the outliers in the 61–120 interval bin, all coming from patient P-5 whose growth pattern is irregular.

\begin{table}[t]
\caption{Prediction performance across different $\Delta \ell_2$ intervals.}
\label{tab:time_intervals}
\centering
\small
\begin{tabular}{@{}lccc@{}}
\toprule
$\Delta \ell_2$ intervals & N° instance & RMSE $\downarrow$ & DICE $\uparrow$ \\ 
\cmidrule(r){1-1} \cmidrule(lr){2-2} \cmidrule(lr){3-3} \cmidrule(l){4-4}
$0 - 60$ & 26 & \textbf{0.042 $\pm$ 0.01} & \textbf{0.781 $\pm$ 0.04} \\
$61 - 120$ & 99 & 0.043 $\pm$ 0.03 & 0.729 $\pm$ 0.08 \\
$121 - 180$ & 59 & 0.055 $\pm$ 0.03 & 0.737 $\pm$ 0.07 \\
$181 - 280$ & 73 & 0.047 $\pm$ 0.02 & 0.737 $\pm$ 0.06 \\
$281 - 365$ & 32 & 0.054 $\pm$ 0.02 & 0.736 $\pm$ 0.03 \\
$366+$ & 11 & 0.062 $\pm$ 0.01 & 0.699 $\pm$ 0.03 \\
\bottomrule
\end{tabular}
\end{table}

Figure \ref{fig:scatter} presents a scatter of predicted versus true tumor areas, with the diagonal identity line indicating perfect agreement. The mean absolute distance from this line is 13.14 $\text{cm}^2$. Excluding patient P-5, whose highly irregular growth accounts for 19 of the 25 largest deviations, reduces this error to 10.80 $\text{cm}^2$. Points more than 30 $\text{cm}^2$ distance from the identity line are marked as outliers with red circles (mostly from subject P-5). Importantly, most predictions fall within 0–15 $\text{cm}^2$ distance from the diagonal, underscoring our model’s consistant overall accuracy in estimating tumor area.

\begin{figure}[t]
    \begin{center}
    \includegraphics[width=\columnwidth]{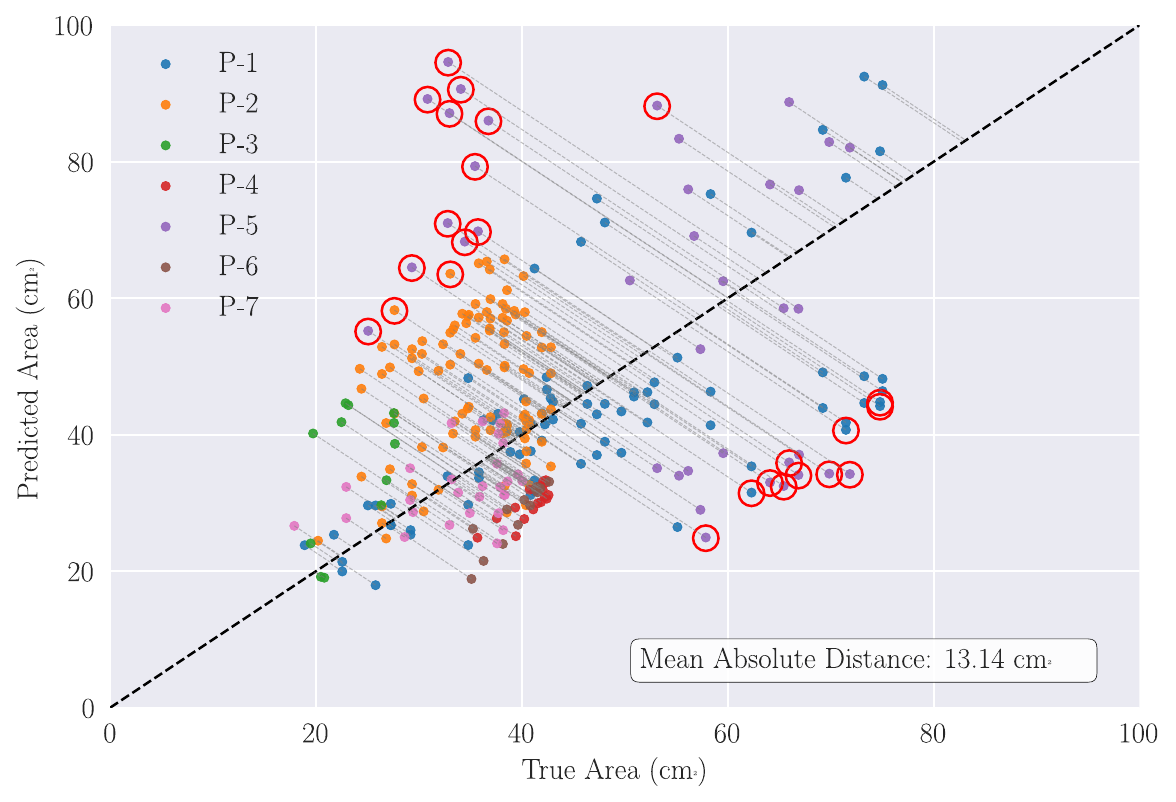}
    \caption{Scatter plot of predicted vs. true tumor areas with the identity line denoting perfect agreement. Mean absolute deviation is 13.14 $\text{cm}^2$, and outliers ($> 30 \;\text{cm}^2$) are highlighted in red.}
    \label{fig:scatter}
    \end{center}
\end{figure}

\begin{figure*}[t]
    \begin{center}
    \includegraphics[width=15.5cm]{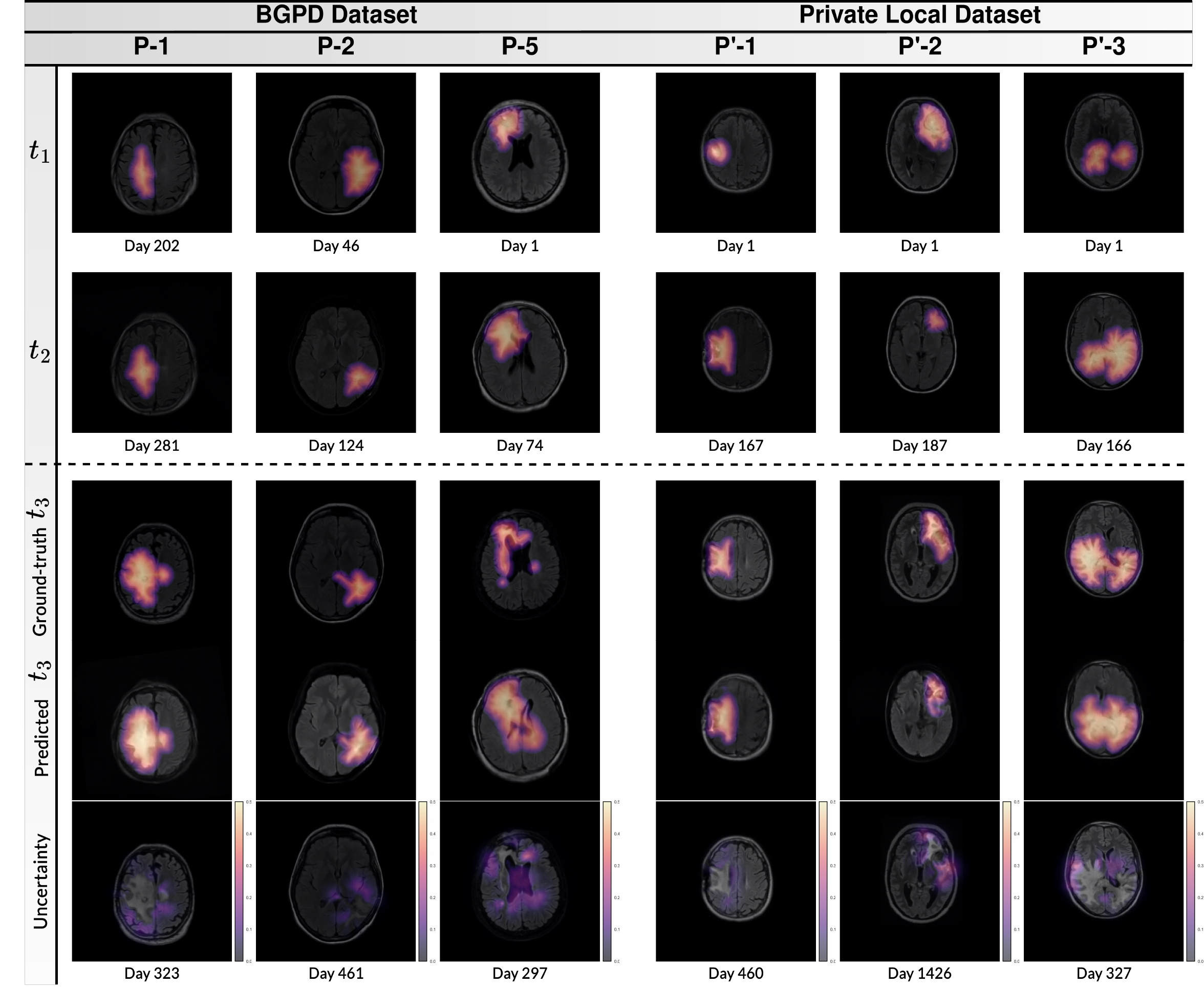}
    \caption{Qualitative predictions for patients P-1, P-2, and P-5 from the BGPD dataset, and P'-1, P'-2, and P'-3 from our local private dataset. Columns $t_1$ and $t_1$ display the source FLAIR MRIs, with acquisition days labeled. Target images include the ground truth and predicted scans overlaid with their respective tumor probability maps, alongside corresponding uncertainty maps. Patient P-5 (bottom row) highlights a challenging case: at $t_3$, the tumor splits into multiple disconnected subregions, resulting in higher prediction error and uncertainty (\ref{sec:results}).}
    \label{fig:burdenko_samples}
    \end{center}
\end{figure*}

\begin{table*}[t!]
\label{tab:per_patient}
\centering
\resizebox{\textwidth}{!}{%
\small
\begin{tabular}{@{}lccccccccc@{}}
\toprule
\multirow{2}{*}{Patient} & \multicolumn{5}{c}{Prediction metrics} & \multicolumn{4}{c}{Generative metrics} \\
\cmidrule(lr){2-6} \cmidrule(l){7-10}
 & RMSE $\downarrow$ & CE $\downarrow$ & KL $\downarrow$ & DICE $\uparrow$ & Pearson's $r \uparrow$ & MSE $\times 10^{-3}$ $\downarrow$ & PSNR $\uparrow$ & SSIM $\uparrow$ & LPIPS $\downarrow$ \\
\cmidrule(r){1-1} \cmidrule(lr){2-2} \cmidrule(lr){3-3} \cmidrule(lr){4-4} \cmidrule(lr){5-5} \cmidrule(lr){6-6} \cmidrule(lr){7-7} \cmidrule(lr){8-8} \cmidrule(lr){9-9} \cmidrule(l){10-10}
P-1 & 0.039 & 0.106 & 0.123 & 0.802 & 0.852 & 0.253 & 26.578 & 0.661 & 0.071\\
P-2 & 0.046 & 0.082 & 0.284 & 0.758 & 0.789 & 0.329 & 26.060 & 0.883 & \textbf{0.051}\\
P-3 & \textbf{0.027} & \textbf{0.070} & 0.153 & 0.676 & 0.829 & \textbf{0.104} & \textbf{29.468} & \textbf{0.904} & 0.064\\
P-4 & 0.045 & 0.121 & \textbf{0.063} & 0.682 & \textbf{0.928} & 0.619 & 22.889 & 0.829 & 0.083\\
P-5 & 0.105 & 0.210 & 0.574 & 0.609 & 0.587 & 0.857 & 21.812 & 0.760 & 0.081\\
P-6 & 0.046 & 0.127 & 0.079 & 0.807 & 0.870 & 0.156 & 28.137 & 0.840 & 0.069\\
P-7 & 0.029 & 0.152 & 0.152 & \textbf{0.821} & 0.888 & 0.374 & 25.024 & 0.785 & 0.072\\
\cmidrule(r){1-1} \cmidrule(lr){2-2} \cmidrule(lr){3-3} \cmidrule(lr){4-4} \cmidrule(lr){5-5} \cmidrule(lr){6-6} \cmidrule(lr){7-7} \cmidrule(lr){8-8} \cmidrule(lr){9-9} \cmidrule(l){10-10}
Avg & 0.048 $\pm$ 0.02 & 0.114 $\pm$ 0.04 & 0.204 $\pm$ 0.16 & 0.736 $\pm$ 0.07 & 0.820 $\pm$ 0.10 & $0.385 \pm 0.24$ & $25.710 \pm 2.52$ & $0.809 \pm 0.07$ & $0.070 \pm 0.01$ \\
\midrule
P'-1 & 0.047 & 0.090 & 0.308 & 0.767 & \textbf{0.886} & 0.100 & 19.983 & 0.758 & \textbf{0.054}\\
P'-2 & 0.037 & \textbf{0.059} & 0.297 & 0.552 & 0.751 & 0.188 & 17.246 & 0.774 & 0.086\\
P'-3 & \textbf{0.035} & 0.114 & \textbf{0.033} & \textbf{0.911} & 0.728 & \textbf{0.015} & \textbf{28.325} & \textbf{0.795} & 0.069\\
P'-4 & 0.057 & 0.136 & 0.067 & 0.825 & 0.808 & 0.080 & 21.030 & 0.735 & \textbf{0.054}\\
P'-5 & 0.061 & 0.133 & 0.090 & 0.742 & 0.839 & 0.120 & 19.187 & 0.690 & 0.097\\
\cmidrule(r){1-1} \cmidrule(lr){2-2} \cmidrule(lr){3-3} \cmidrule(lr){4-4} \cmidrule(lr){5-5} \cmidrule(lr){6-6} \cmidrule(lr){7-7} \cmidrule(lr){8-8} \cmidrule(lr){9-9} \cmidrule(l){10-10}
Avg & 0.048 $\pm$ 0.01 & 0.106 $\pm$ 0.03 & 0.159 $\pm$ 0.12 & 0.759 $\pm$ 0.12 & $0.802 \pm 0.05$ & $0.101 \pm 0.05$ & $21.154 \pm 3.79$ & $0.751 \pm 0.03$ & $0.072 \pm 0.02$ \\
\bottomrule
\end{tabular}
}
\caption{Predictive and generative results of each patient using our proposed method. We also report the Pearson's correlation scores between predicted and ground truth probabilistic GTV maps.}
\end{table*}

\begin{figure}[t]
    \begin{center}
    \includegraphics[width=\columnwidth]{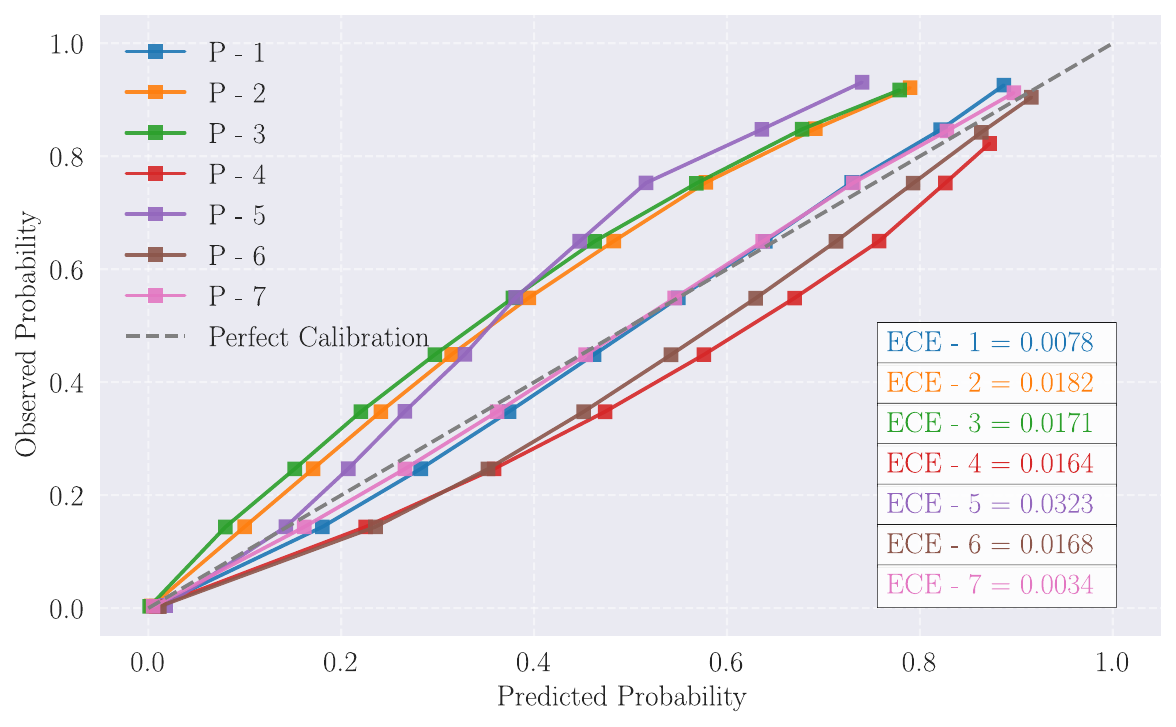}
    \caption{ Calibration curves for each test patient showing predicted probability versus observed tumor frequency across probability bins.}
    \label{fig:calibration_per_patient}
    \end{center}
\end{figure}

Table \ref{tab:per_patient} reports prediction and generative metrics for each of the seven test patients. We see substantial inter‐patient variability driven largely by tumor evolution and scan intervals. For example, P-3 has the lowest RMSE (0.027), CE (0.070), and MSE (0.104 $\times 10^{-3}$), along with the highest PSNR (29.47) and SSIM (0.904), reflecting its relatively simple local evolving tumors that the model captures with high fidelity. In contrast, P-5 exhibits the poorest performance (DSC = 0.609), underscoring the challenge of its complex, erratic growth pattern. Calibration curves for P-5 (see Figure \ref{fig:calibration_per_patient}) are similarly misaligned, with the highest ECE (0.0323), and sample visualizations in Figure \ref{fig:burdenko_samples} highlights the tumor’s fragmentation across subregions at target $t_3$ timepoint, which can explain the drop in its results. This atypical progression illustrates the difficulty of training predictive models under data-scarce regimes with limited patterns of change. Patients with moderate evolution (P-1, P-2, P-4, P-6, P-7) fall between these extremes, with RMSE values of 0.039–0.046 and DSC scores of 0.758–0.821, assessing the model’s reliability when tumor evolution follows more predictable trajectories, while also capturing more heterogeneous growth patterns (P-1 and P-2 in Figure \ref{fig:burdenko_samples}). This proves the potential of our method to capture more complex long range trajectories even with limited training sets. Furthermore, we also evaluate spatial correlation via Pearson’s score between predicted and ground truth probabilistic GTV maps (in Table \ref{tab:per_patient}). We drop low-density regions, restricting the analysis to tumor pixel and its surrounding only ($P > 0.2$). Most patients exhibit strong correlations ($\approx0.8–0.9$), reflecting accurate spatial tracking of tumor morphology, whereas P-5 shows again a markedly lower value, consistent with its poor RMSE/DSC and erratic growth pattern. This alignment between correlation and our other metrics further underscores the model's consistent predictions.

We further evaluated the generalizability of our model, which was exclusively trained on the BGPD cohort without any external pretraining, by assessing its performance on an independent, private test set of five patients (P'-1 through P'-5). These external cases faithfully reproduce the predictive trends observed in BGPD, with similar RMSE and DSC ranges, except for patient P'-2, who exhibits a lower DSC yet still delivers decent performance on the other metrics, likely due to a slight shift around the $P_t \approx 0.8$ GTV boundary probabilities used to compute the DSC metric. The generative metrics likewise remain consistent, with only a modest drop in average SSIM. We observed a perceptual decline and subtle quality differences in some of our private-set predictions compared to BGPD, most clearly for patient P'-3 (see Figure \ref{fig:burdenko_samples}). These discrepancies can be explained by variations in contrast, noise profiles, and the overall different scan quality of our private dataset, which introduce unfamiliar patterns for the generative branch. Nevertheless, the results confirm the robustness of our model in its prediction task across diverse clinical sources. Notably, P'-3 achieves an exceptional DSC of 0.911, demonstrating promising prediction accuracy despite tumor changes between source and target timepoints, further assessing our model’s generalizability. Given the lack of similar studies, direct comparisons with the same data are not possible, however, the work in \cite{liu2025treatment} is the only work using comparable data to our knowledge, with a reported DSC of 0.719 despite using an extra input scan, underscoring our approach’s strong performance.

\begin{figure}[t]
    \begin{center}
    \includegraphics[width=8cm]{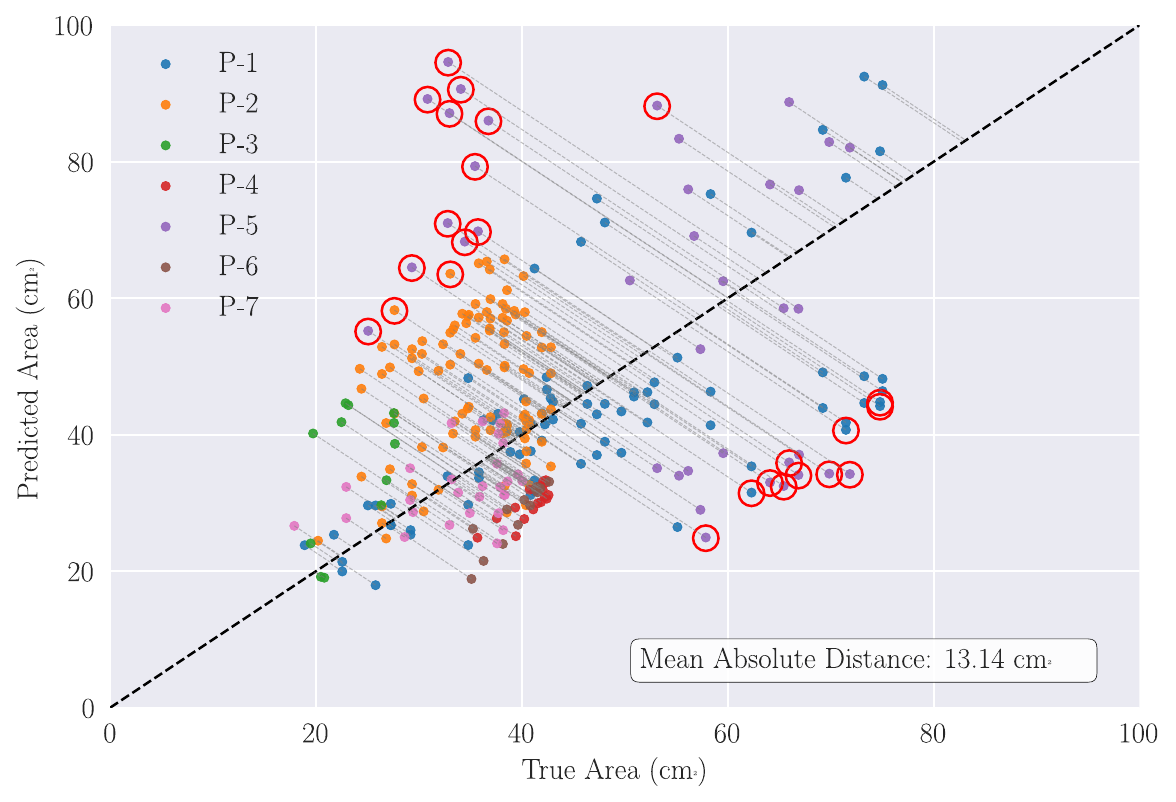}
    \caption{Scatter plot of predicted vs. true tumor areas with the identity line denoting perfect agreement. Mean absolute deviation is 13.14 $\text{cm}^2$, and outliers ($> 30 \;\text{cm}^2$) are highlighted in red.}
    \label{fig:scatter}
    \end{center}
\end{figure}

\section{Conclusion}
In this work, we introduced a novel architecture GliomaDiff, a multi‐task diffusion framework for time‐agnostic prediction of glioma progression from longitudinal MRI. By reformulating tumor masks as logistic‐normalized signed distance fields, we enabled continuous uncertainty quantification and improved alignment with ground truth tumor evolution in pixel-level. A dedicated tumor evolution module learned dense inter‐scan vector fields to capture non‐linear tumor evolution dynamics, and a targeted data‐augmentation pipeline, combining both standard augmentations and synthesized follow‐up sequences, helped addressing the scarcity of longitudinal data. 
Our method achieves reliable forecasts up to six months ahead, with calibrated uncertainty estimates that remain coherent across diverse patient dynamics. 

\newpage






\bibstyle{elsarticle-harv} 
\bibliography{main}





\end{document}